\pdfoutput=1

\documentclass[11pt]{article}

\usepackage[preprint]{acl}

\usepackage{multirow}
\usepackage{placeins}
\usepackage{times}
\usepackage{latexsym}
\usepackage{pbox}
\usepackage[T1]{fontenc}

\usepackage[utf8]{inputenc}

\usepackage{microtype}

\usepackage{inconsolata}

\usepackage{colortbl}
\usepackage{graphicx}
\usepackage{xspace}
\usepackage{tabularx}
\usepackage{tabulary}
\usepackage{amsmath}
\usepackage{booktabs}
\usepackage{listings}
\usepackage[most]{tcolorbox}
\usepackage{colortbl}
\usepackage{cuted}
\usepackage{xcolor}
\usepackage{kotex}
\usepackage{pifont}
\usepackage{dirtytalk}
\usepackage[utf8]{inputenc}
\usepackage[T1]{fontenc}
\usepackage[perpage]{footmisc}
\usepackage{hyperref}
\usepackage{colortbl}

\newtcolorbox{instructionsbox}[1][]{
  colframe=cyan!75!black,    
  colback=green!5!white,     
  coltitle=black,            
  title=#1,                  
  rounded corners,           
  boxrule=0.5mm,             
  boxsep=5pt,                
  toptitle=1mm,              
  bottomtitle=1mm,           
  left=0pt,                 
  right=0pt,                
  top=0pt,                   
  bottom=0pt,                
  fonttitle=\bfseries        
}

\newcommand{\dataset}{\textsc{HRM8K}\xspace}
\newcommand{\traindataset}{\textsc{UST}\xspace}
\newcommand{\method}{\textsc{UST}\xspace}

\title{Understand, Solve and Translate:\\Bridging the Multilingual Mathematical Reasoning Gap}

\author{Hyunwoo Ko{\textsuperscript{1\footnotemark[1]}} \quad Guijin Son{\textsuperscript{1,2\thanks{Equal Contribution}}} \quad \textbf{Dasol Choi}{\textsuperscript{2}\footnotemark[1]} \\ \\ 
OneLineAI{\textsuperscript{1}} \quad Yonsei University{\textsuperscript{2}} \\
\texttt{hyunwooko@onelineai.com} \quad \texttt{spthsrbwls123@yonsei.ac.kr} \\ 
}

\begin{document}
\maketitle
\begin{abstract}
Large language models (LLMs) demonstrate exceptional performance on complex reasoning tasks. However, despite their strong reasoning capabilities in high-resource languages (e.g., English and Chinese), a significant performance gap persists in other languages. To investigate this gap in Korean, we introduce \dataset, a benchmark comprising 8,011 English-Korean parallel bilingual math problems. 
Through systematic analysis of model behaviors, we identify a key finding: these performance disparities stem primarily from difficulties in comprehending non-English inputs, rather than limitations in reasoning capabilities. 
Based on these findings, we propose \method (Understand, Solve, and Translate), a method that strategically uses English as an anchor for reasoning and solution generation. By fine-tuning the model on 130k synthetically generated data points, \method achieves a 10.91\% improvement on the \dataset benchmark and reduces the multilingual performance gap from 11.6\% to 0.7\%. 
Additionally, we show that improvements from \method generalize effectively to different Korean domains, demonstrating that capabilities acquired from machine-verifiable content can be generalized to other areas. We publicly release the benchmark, training dataset, and models\footnote{\url{https://huggingface.co/datasets/HAERAE-HUB/HRM8K}}.

\end{abstract}

\section{Introduction}

Large language models (LLMs) have made remarkable progress in reasoning tasks, often surpassing expert human performance~\citep{OpenAI2024o1, Anthropic2024}. However, this exceptional reasoning capability is primarily observed in high-resource languages, with significant performance gaps in lower-resource languages~\citep{huang2023not, li2024quantifying}. This disparity likely stems from LLMs' difficulty in transferring their foundational capabilities, including reasoning skills learned in high-resource languages like English or Chinese, to lower-resource languages~\citep{chen2023breaking, dubey2024llama}.

To investigate this gap in Korean mathematical reasoning, we introduce \dataset, a bilingual benchmark comprising 8,011 questions in both Korean and English. The questions are carefully curated from existing benchmarks~\citep{cobbe2021training, hendrycks2021measuring, gao2024omni} and Korean examinations to create a perfectly parallel evaluation structure. Through systematic evaluation on \dataset, we reveal that the performance gap primarily stems from difficulties in comprehending non-English inputs, rather than limitations in reasoning capabilities. This finding challenges previous studies that suggest using English chain-of-thought (CoT) reasoning for multilingual questions~\citep{shi2022language}, as we show that LLMs are heavily influenced by the input language itself.

Based on these insights, we propose \method (Understand, Solve, and Translate), a training method that strategically uses English as an anchor for reasoning and solution generation. Our approach builds on recent findings that LLMs effectively use English as a pivot language for processing multilingual inputs~\citep{zhong2024beyond}. By training on 130k synthetically generated instances, \method achieves a 10.91\% improvement on the \dataset benchmark and reduces the multilingual performance gap from 11.6\% to 0.7\%. Furthermore, we demonstrate that these improvements generalize beyond mathematics to different Korean domains, suggesting broader applications.

In summary, the main contributions of this work are as follows:
\begin{itemize}
    \item We identify through systematic analysis that multilingual performance gaps primarily stem from input comprehension difficulties rather than reasoning limitations.
    \item We propose \method, a training method that effectively leverages English reasoning capabilities for non-English inputs, demonstrating significant performance improvements.
    \item We introduce \dataset, the first Korean mathematics reasoning benchmark with 8,011 parallel questions, enabling systematic evaluation of multilingual reasoning capabilities.
\end{itemize}

\section{Related Work}

\paragraph{Mathematics Benchmarks}

Mathematical reasoning has emerged as a crucial capability for language models~\citep{OpenAI2024, qwq2024, zhao2024marco}, leading to the development of numerous benchmarks and datasets~\citep{ling2017program, amini2019mathqa, patel2021nlp, saxton2019analysing}. 
Traditional datasets such as GSM8K~\citep{cobbe2021training} and MATH~\citep{hendrycks2021measuring} primarily target grade-school to undergraduate-level problems, while more recent efforts introduce Olympiad-level challenges~\citep{zheng2021minif2f, he2024olympiadbench, huang2024olympicarena, fang2024mathodyssey, gao2024omni}.

Although these benchmarks prove valuable for evaluating English-language mathematical reasoning, fewer resources exist for non-English or bilingual math problems~\citep{shi2022language, chen2023breaking, wu2024conceptmath}. In the Korean context, most benchmarks emphasize language understanding~\citep{park2021klue, son2023removing}, general knowledge~\citep{son2023hae, kim2024click}, or commonsense reasoning~\citep{son2024llm, son2024krx}, with mathematics being largely underrepresented. While the Open Ko-LLM Leaderboard~\citep{park2024open} has begun translating some popular English benchmarks into Korean, those translated sets are not publicly accessible. Meanwhile, KMMLU~\citep{son2024kmmlu} includes only about 100 math problems, insufficient for broader evaluation. To address this gap, we propose \dataset, a large-scale bilingual Korean-English math benchmark comprising 8{,}011 problems, covering both competition-level Korean questions and parallel translations of existing English benchmarks.

\paragraph{Multilingual Reasoning and Language Models}

Recent LLMs have shown remarkable performance in English~\citep{OpenAI2024o1, Anthropic2024, touvron2023llama2}, but many still underperform in multilingual scenarios~\citep{lai2024llms, dubey2024llama}. Such performance discrepancies are attributed to limited exposure to lower-resource languages during pre-training. Consequently, much research has focused on enhancing multilingual reasoning skills, including methods that explicitly use English as a `pivot’ for cross-lingual tasks~\citep{zhao2024large, zhu2024question}. For example, PLUG~\citep{zhang2023plug} aligns internal reasoning in different languages, enabling the model to leverage stronger English reasoning for other languages.

Despite these developments, few works have thoroughly examined how to best calibrate reasoning between high- and low-resource languages in \emph{complex mathematical} contexts. Some studies investigate altering the fraction of multilingual data in training~\citep{anonymous2024enhancing} or conduct smaller-scale experiments on bilingual math tasks~\citep{shi2022language}, yet a clear, large-scale solution remains elusive. Against this backdrop, our work introduces \method, a multilingual reasoning method that intentionally routes math problems in lower-resource languages through English-based reasoning. We show that this strategy drastically narrows the performance gap and advances multilingual math capabilities.

\begin{table*}[ht]
\centering
\fontsize{9}{11}\selectfont
\begin{tabular}{c@{\hspace{2em}}l@{\hspace{1.0em}}c@{\hspace{1.5em}}l}
\toprule
\addlinespace[0.5em]
\textbf{Category} & \textbf{Subset} & \textbf{\# of Instances} & \multicolumn{1}{c}{\textbf{Short Description}}
\\ \midrule
\addlinespace[0.3em]
KSM & KMO & 730 & \parbox{0.55\textwidth}{ Mathematics competition for high school students in South Korea, top-performers are selected as representatives for the IMO.~\citep{kmo_kmo}} \\
\addlinespace[0.3em]
\textbf{1.4K Total} & KJMO & 62 & \parbox{0.55\textwidth}{KMO for junior students, up to age 13~\citep{kjmo_kjmo}} \\
\addlinespace[0.3em]
 & CSAT & 210 & \parbox{0.55\textwidth}{Questions from the Korean national university entrance exam and official mock exams, we only filter questions that have an error rate exceeding 70\%.~\citep{csat_csat}} \\
 \addlinespace[0.3em]
 & KMS & 82 & \parbox{0.55\textwidth}{Math olympiad for university students, organized by the  Korean Mathematical Society~\citep{kms_kms}} \\
 \addlinespace[0.3em]
 & TQ & 344 & Question from the national assessment test for math teacher certification~\citep{tq_tq} \\ 
\addlinespace[0.3em]
\midrule
\addlinespace[0.3em] 
Prior Sets & GSM8K & 1,319 & \parbox{0.55\textwidth}{Grade school math word problems created by human problem writers~\citep{cobbe2021training}} \\
\addlinespace[0.3em]
\textbf{6.5K Total} & MATH & 2,885 & \parbox{0.55\textwidth}{Competition-level mathematics problems. We only include questions with numeric answers~\citep{hendrycks2021measuring}} \\
\addlinespace[0.3em]
 & Omni-MATH & 1,909 & \parbox{0.55\textwidth}{Olympiad-level problems collected from international and Chinese math competitions. We only include questions with numeric answers~\citep{gao2024omni}} \\ 
 \addlinespace[0.3em]
 & MMMLU & 470 & \parbox{0.55\textwidth}{The MMLU~\citep{hendrycks2020measuring} dataset translated by professional human translators~\citep{openai_mmmlu}} \\
\addlinespace[0.3em]
\bottomrule
\end{tabular}
\caption{Summary of dataset sources used in \dataset}
\label{tab:dataset_summary}
\end{table*}
\section{\dataset}

In this section, we introduce the composition of the \dataset benchmark and explain its construction process. We also conduct a contamination check to ensure data quality. Detailed information about each dataset and post-processing methods are provided in Appendix~\ref{app:detail_on_dataset}.

\subsection{Benchmark Formulation}\label{sec_bencmark_formulation}

The \dataset benchmark is a bilingual math dataset that consists of two major subsets: \textbf{Korean School Math} (KSM) and \textbf{Prior Sets}. Each subset is available in both Korean and English (see Table~\ref{tab:dataset_summary} for details).

\paragraph{KSM}~This subset contains 1,428 challenging math problems sourced from Korean Olympiads and competition-level exams, irrespective of the target age group. As a result, even questions originally intended for younger students still require substantial reasoning ability to solve. To collect these questions, the authors manually captured screenshots and applied GPT-4o’s OCR to convert the text, followed by a thorough validity check. (See Appendix~\ref{app:prompt_template} for the OCR prompt.)

\paragraph{Prior Sets}~This subset comprises 6,583 problems drawn from established English math benchmarks, including GSM8K~\citep{cobbe2021training}, MATH~\citep{hendrycks2021measuring}, Omni-MATH~\citep{gao2024omni}, and MMMLU~\citep{openai_mmmlu}. To streamline translation and evaluation, we include only instances with numeric answers, excluding problems that require text-based, proof-oriented, or equation-based final answers. In particular, proof-type questions would necessitate a more complex LLM-as-a-Judge approach~\citep{zheng2023judging, shi2024judging, park2024offsetbias} instead of simpler machine verification. Lastly, from the MMMLU dataset, we select only three math-related subsets: \texttt{abstract\_algebra}, \texttt{college\_mathematics}, and \texttt{high\_school\_mathematics}. Questions for the MMMLU dataset is multiple-choice question answering format.

\subsection{Benchmark Construction}\label{sec_construction}

We translate all instances in both subsets into English and Korean using GPT-4o~\citep{OpenAI2024}, and then conduct human review to remove any inaccurate translations. This fully parallel design enables a more direct analysis of multilingual performance gaps. Furthermore, because KSM is translated from Korean to English and the other benchmarks are translated from English to Korean, we avoid depending solely on a single translation direction. This bidirectional approach also helps detect translation artifacts: if a particular pattern appears only in one direction, it may be due to translation-related issues rather than the inherent difficulty of the questions. For further details on the dataset construction, please refer to Appendix~\ref{app:detail_on_dataset}.

\subsection{Contamination Check}\label{sec_contamination}

Benchmark contamination, where evaluation questions appear in the model’s pretraining data, is increasingly recognized as a key concern in LLM evaluations~\citep{deng2023investigating, roberts2023cutoff}. Large-scale internet-crawled corpora~\citep{gao2020pile, weber2024redpajama} raise the likelihood of model memorization, potentially leading to inflated performance on evaluation benchmarks~\citep{zhang2024careful, zhao2024mmlu}.
Ensuring that a newly proposed benchmark is entirely uncontaminated is nearly impossible, as many companies do not disclose the specifics of their pretraining mixtures~\citep{aryabumi2024aya, mishra2024granite}, and logit-based detection methods are not yet well-established~\citep{xu2024benchmarking}. In this work, we make our best effort to verify whether the dataset is included in publicly available large-scale Korean corpora. The contamination check is focused on the KSM subset, the only subset that was crawled in this work.

To ensure that the KSM subset is not present in common pretraining corpora, we perform a contamination check against FineWeb-2~\citep{penedo2024fineweb-2}, the biggest Korean corpora available. This dataset contains 58 million Korean documents, totaling 95\,GB, collected by the CommonCrawl foundation (2013--2024). We first identify 149 documents that match the external sources used to compile \dataset. Then, we search these documents for exact string matches from KSM’s questions; no matches were found. We hypothesize that this absence arises because the authors manually downloaded PDF or HWP files and selectively extracted questions, making them unlikely to appear in standard web crawls. Consequently, we conclude that the KSM problems are highly unlikely to have been seen during the LLMs’ pretraining phase.

\section{Multilingual Performance Gaps}\label{sec_mpg}

A recurring observation in large language models (LLMs) is that performance can vary significantly depending on the language of the prompt, even if the underlying task remains the same. We confirm this phenomenon on the \dataset benchmark: as shown in Table~\ref{tab_hrm8k_results}, simply changing both the input and reasoning language from Korean to English yields an 11\% performance boost, suggesting a notable gap in multilingual reasoning.

This section further investigates the causes of this gap. We first describe our experimental design (Section~\ref{sec_experiment_design}), then analyze the results (Section~\ref{sec_evaluatio_results}), and finally explore how multi-step prompting might mitigate these issues (Section~\ref{sec_multi_task}).

\subsection{Experimental Design}\label{sec_experiment_design}

Let a model’s final performance $P$ be determined by two factors: the language of the \textit{input} ($L_\text{input}$) and the language used for \textit{reasoning} ($L_\text{reason}$). Formally,
\[
  P = f(L_\text{input}, L_\text{reason}).
\]
In the context of solving Korean math problems, there are two key requirements:
\begin{description}
    \item[Comprehension:] The model must understand the question, which is provided in Korean:
    \[
    P \propto \text{Comp.}(L_\text{input} = \text{Korean}).
    \]
    \item[Reasoning:] It must also perform the reasoning steps in Korean:
    \[
    P \propto \text{Reasoning}(L_\text{reason} = \text{Korean}).
    \]
\end{description}
To examine which factor is more critical, we evaluate three cross-lingual setups: (1) Korean-to-Korean (K2K), (2) Korean-to-English (K2E), and (3) English-to-English (E2E). We exclude the English-to-Korean (E2K) scenario because models typically fail to maintain a Korean chain-of-thought when the input is given in English. Further details, including the prompts used, can be found in Appendices~\ref{app:detail_on_exp} and~\ref{app:prompt_template}.

\subsection{Evaluation Results}\label{sec_evaluatio_results}

\begin{table}[ht]
\centering
\fontsize{10}{11}\selectfont
\begin{tabular}{cc|ccc}
\toprule
\multicolumn{2}{c}{\textbf{Prompting Type}} & \textbf{K2K} & \textbf{K2E} & \textbf{E2E} \\
\multirow{2}{*}{Language} & \(L_\text{input}\) & Ko & Ko & En \\
 & \(L_\text{reason}\) & Ko & En & En \\
 \midrule
\multirow{3}{*}{Qwen2.5} & 1.5B & 16 & 21 {\color[HTML]{6AA84F} (+5)} & 37 {\color[HTML]{6AA84F} (+21)} \\
 & 7B & 40 & 41 {\color[HTML]{6AA84F} (+1)} & 51 {\color[HTML]{6AA84F} (+11)} \\
 & 72B & 58 & 60 {\color[HTML]{6AA84F} (+2)} & 63 {\color[HTML]{6AA84F} (+5)} \\
 \midrule
\multirow{3}{*}{Llama3.1/2} & 1B & 7 & 7 (0) & 22 {\color[HTML]{6AA84F} (+15)} \\
 & 8B & 28 & 26 {\color[HTML]{E06666} (-2)} & 39 {\color[HTML]{6AA84F} (+11)} \\
 & 70B & 45 & 45 (0) & 55 {\color[HTML]{6AA84F} (+12)} \\
 \midrule
\multicolumn{3}{c}{Average Delta} & \cellcolor[HTML]{D9EAD3} +1 & \cellcolor[HTML]{B6D7A8} +11 \\
 \bottomrule
\end{tabular}
\caption{Performance of Qwen2.5 and Llama3.1/2 models on the \dataset benchmark depending on the input and reasoning language. Number in bracket denote its gain compared to the K2K prompt. }
\label{tab_hrm8k_results}
\end{table}

Table~\ref{tab_hrm8k_results} summarizes the outcomes for each configuration. Overall, performance tends to increase with model size, and larger models show smaller gaps across languages. We highlight two findings:

\paragraph{Effect of Input  Language} 
Switching from Korean input (K2E) to an entirely English setup (E2E) yields an average improvement of 11\%. In particular, Qwen2.5-7B and Llama-3.1-8B drop by 10\% and 13\%, respectively, when forced to process Korean input. This underscores the significance of \(L_\text{input}\) in model performance.

\paragraph{Effect of Reasoning Language} 
In contrast, comparing K2K to K2E shows an average difference of only 1\%, suggesting that the language of the \emph{reasoning process} has a relatively small impact once the model has already ingested Korean input. Simply allowing the model to produce its chain-of-thought in English does not fully recover performance lost from reading a Korean prompt.

In short, enabling English-based reasoning alone is insufficient to close the multilingual gap. Instead, the limiting factor appears to be how well the model can \emph{comprehend} Korean inputs.

\begin{figure*}[h]
    \centering
    \includegraphics[width=\linewidth]{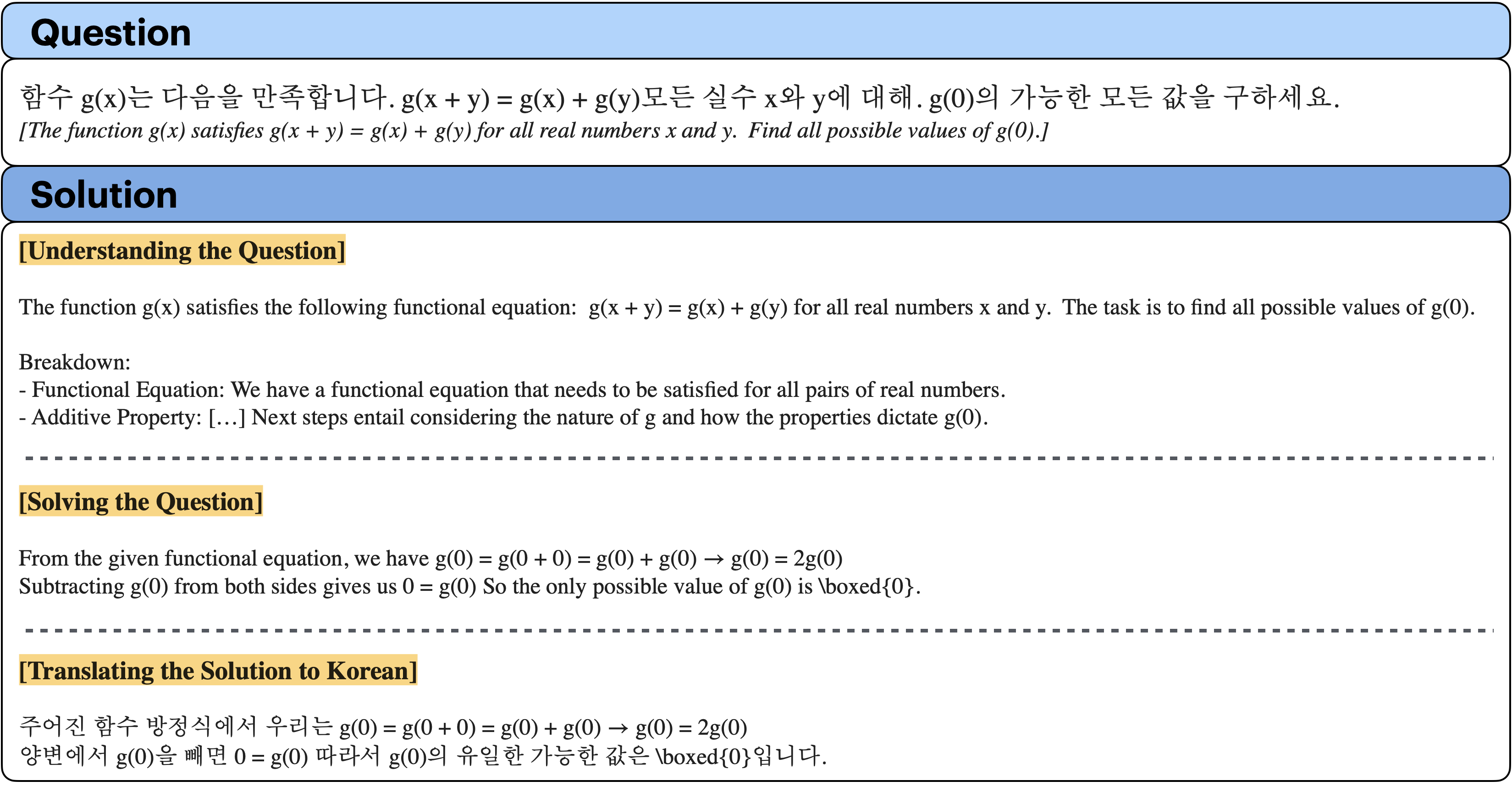}
    \caption{Example of \method process. When presented with a problem in Korean, the model generates Korean answers through the following processes highlighted in yellow: \textit{Understanding the Question}, \textit{Solving the Question}, and \textit{Translating the Solution into Korean}.}
    \label{fig:tas_example}
\end{figure*}

\subsection{Multi-Step Inference}\label{sec_multi_task}

\begin{figure*}[h]
    \centering
    \includegraphics[width=\linewidth]{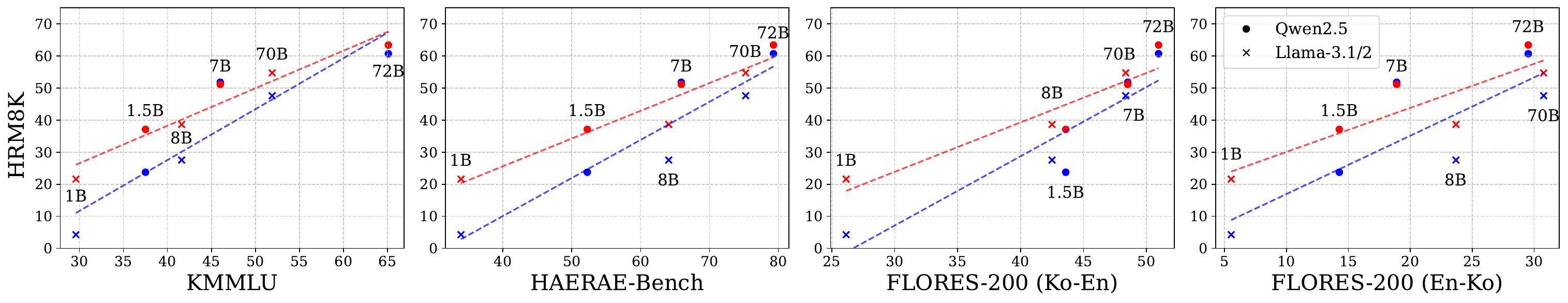}
    \caption{Comparison of \dataset performance (vertical axis) and three additional benchmarks (KMMLU, HAERAE-Bench, FLORES-200) for Qwen2.5 and Llama-3.1/2 across different model sizes. TE2E (blue) translates Korean input to English before solving; E2E (red) uses an English prompt from the start.}
    \label{fig:kmmlu_hr_result}
\end{figure*}

A straightforward approach to alleviating these comprehension issues is to split the task into multiple inference steps, where the model explicitly translates the Korean question into English first. We denote this as \emph{Translated-English} (TE) in the first step, followed by an English-to-English (E2E) reasoning step in the second inference. This overall pipeline, called TE2E, uses a five-shot prompt for the translation stage.

In Figure~\ref{fig:kmmlu_hr_result}, we compare TE2E (blue) and E2E (red) on three additional benchmarks: KMMLU~\citep{son2024kmmlu}, HAERAE-Bench~\citep{son2023hae}, and FLORES~\citep{nllb-24}. For larger models with stronger Korean proficiency, TE2E and E2E become more similar, implying that multi-step inference can indeed help if the model already translates Korean accurately. However, for smaller models with weaker Korean skills, the performance gap remains pronounced. (Further details on the TE prompt are provided in Appendix~\ref{app:prompt_template}.)

\section{Understand, Solve and Translate}\label{sec_methods}

Based on the experiments in Section~\ref{sec_mpg}, we conjecture that for smaller models improving their ability to understand Korean questions and allowing it to reason in English can address or bypass the constraints limiting LLMs in solving Korean questions. To validate this hypothesis, we fine-tune LLMs on a custom dataset designed to guide models through three stages: understanding Korean questions, solving them in English, and translating solutions back to Korean.

In this section, we explain the details of our training dataset (Section~\ref{sec_ust_construction}), report performance gains (Section~\ref{sec_ust_performance}), and conduct ablations on the effectiveness of the training (Section~\ref{sec_ust_ablation}).

\begin{table*}[ht]
\centering
\fontsize{11}{13}\selectfont
\begin{tabular}{l|ccccc|c}
\toprule
\multicolumn{1}{l}{\textbf{Models}} & \textbf{GSM8K} & \textbf{MATH} & \textbf{OMNI\_MATH} & \textbf{MMMLU} & \textbf{KSM} & \textbf{Avg.} \\
\midrule
\multicolumn{7}{l}{\textit{Proprietary or Large  Models}} \\
\midrule
GPT-4o & 91.21 & 74.45 & 30.75 & 68.72 & 22.83 & 57.59 \\
GPT-4o-Mini & 87.57 & 70.68 & 26.45 & 63.40 & 19.40 & 53.50 \\
Qwen2.5-72B-Instruct & 90.07 & 72.06 & 30.96 & 66.60 & 23.46 & 56.63 \\
Llama-3.1-70B-Instruct & 79.08 & 56.05 & 19.85 & 60.00 & 13.10 & 45.61 \\
\midrule
\multicolumn{7}{l}{\textit{Qwen2.5-7B-Instruct}} \\
\midrule
K2K-Prompting & 66.41 & 50.36 & 18.96 & 50.00 & 11.83 & 39.52 \\
K2E-Prompting & 65.20 & 54.59 & 20.22 & 49.79 & 16.67 & 41.29 \\
E2E-Prompting & 81.35 & 68.87 & 27.29 & 57.02 & 21.08 & 51.12 \\
Ours & 80.06 & 68.53 & 27.19 & 57.23 & 19.12 & 50.43 \\ 
\bottomrule
\end{tabular}
\caption{Evaluation results on \dataset, comparing large-scale models (top) with different prompting strategies for Qwen2.5-7B-Instruct (bottom). Our UST-trained model achieves comparable performance to E2E-Prompting.}
\label{tab:method_performance}
\end{table*}

\subsection{Training Dataset Construction}\label{sec_ust_construction}
We create the \traindataset (Understand, Solve, and Translate) dataset designed for training multilingual mathematical reasoning. Using GPT-4o-Mini, we generate cross-lingual Chain-of-Thought (CoT) examples that consist of three stages: (1) English Understanding Stage: Breaking down Korean questions and explaining their context and objectives in English; (2) English Solution Stage: Solving the mathematical problem in English; (3) Korean Solution Stage: Translating the English solution back to Korean.
The dataset construction process follows these specific steps:

\paragraph{Step 1: Seed data collection}~We source our initial data from two datasets: OpenMathInstruct-2~\citep{toshniwal2024openmath2} and NuminaMath-CoT~\citep{numina_math_datasets}. From OpenMathInstruct-2's 14 million instruction samples, we randomly select 5 million instances and translate them to Korean using GPT-4o-Mini. For quality control, we use Qwen2.5-Math-RM-72B~\citep{yang2024qwen25mathtechnicalreportmathematical}, a reward model specialized for evaluating the quality of math question and output pairs, to score these instances and retain the top 50,000.

~However, since OpenMathInstruct-2 primarily contains MATH and GSM8K augmentations, it lacks sufficient olympiad-level problems. To address this limitation, we supplement our dataset with samples from NuminaMath-CoT, specifically selecting problems from Aops Forum, AMC, Synthetic AMC, AIME, and Olympiads, and apply the same processing steps.

\paragraph{Step 2: Generating the Understanding Stage}~Our previous experiments show that models perform better on Korean questions when translation and reasoning are separated into distinct steps. This suggests that traditional training methods, which emphasize immediate reasoning, may limit models' ability to process non-English inputs effectively. Such observation aligns with \citet{zhu2024question}'s finding that translation-specific training enhances multilingual reasoning capabilities.
Based on these insights and recent advances in longer CoT generation, we introduce an \textbf{Understanding Stage} to our training pipeline. In this stage, we use GPT-4o-Mini to create structured breakdowns of Korean questions in English, providing both the original question and its solution to ensure alignment between understanding and problem-solving.

\paragraph{Step 3: Generating the Korean Solution Stage}~Our seed datasets originally include an English solution for each sample. In this step, we translate each solution into Korean. At every generation stage, we include prompts to discard samples with incorrect translations or solutions. The final version of the dataset contains approximately 130k samples. Further details of the prompt used are available in Appendix~\ref{app:prompt_template}.

\begin{table}[h]
\centering
\fontsize{11}{12}\selectfont
\begin{tabular}{lccc}
\toprule
 & \textbf{HRM8K} & \textbf{ELO} & \textbf{Token Consum.} \\
\midrule
K2K & 39.52 & 807 & 2,202 \\
\method & 50.43 & 1145 & 7,854 \\
M.S.I & 51.78 & 978 & 11,764 \\
\bottomrule
\end{tabular}
\caption{Comparison of K2K prompting, \method model (ours), and Multi-Step Inference (M.S.I). For more details on M.S.I see Appendix~\ref{app:msi}.}
\label{tab_elo}
\end{table}

\begin{table*}[ht]
\centering
\fontsize{11}{12}\selectfont
\begin{tabular}{l@{\hspace{2em}}c@{\hspace{2em}}c}
\toprule
\multirow{2}{*}{\textbf{Model Configuration}} & \textbf{Stage Language} & \multirow{2}{*}{\textbf{Accuracy (\%)}} \\
& \textbf{(Understand / Solve)} & \\
\midrule
Baseline (K2K) & Korean / Korean & 15.95 \\
Cross-lingual (K2E) & Korean / English & 21.15 \\
English-only (E2E) & English / English & 37.10 \\
\midrule
\multicolumn{3}{c}{\textbf{Ablation Studies}} \\
\midrule
No Understanding & - / Korean & 36.49 \\
Korean Understanding & Korean / Korean & 34.82 \\
No Understanding & - / English & 43.30 \\
Full \method & English / English & \textbf{44.43} \\ 
\bottomrule
\end{tabular}
\caption{Ablation studies on different configurations of \method. The top section shows baseline prompting results, while the bottom section examines the impact of different language settings for each stage.}
\label{tab:ablation_training}
\end{table*}

\subsection{Fine-Tuning with \method}
We fine-tune Qwen2.5-7B using a standard autoregressive objective to generate all three stages (understanding, solving, and translating) in a single inference. Special tokens are inserted between stages to enable selective generation during inference. The model architecture remains unchanged, without parameter freezing or additional parameters.
The training process runs for 3 epochs (approximately 11 hours) on four H100 80GB HBM3 GPUs using DeepSpeed ZeRO-1 parallelism~\citep{rajbhandari2020zero}. Detailed training configurations and hyperparameters are provided in Appendix~\ref{app:finetuning_details}.

\subsection{Performance Analysis}\label{sec_ust_performance}

\paragraph{Effects of Targeted Training}~Table~\ref{tab:method_performance} shows the performance of our Qwen2.5-7B model trained on the \traindataset dataset. Our model achieves higher accuracy (50.43\%) than both baseline approaches: K2K (39.52\%) and K2E (41.29\%) prompting, highlighting the effectiveness of targeted training for English reasoning. Furthermore, this performance is comparable to E2E prompting (51.12\%), suggesting that our model successfully recovers the capabilities observed under ideal conditions where both questions and reasoning are in English.

\paragraph{Effects of Single-Pass Translation}~Instances in the \traindataset dataset integrate translation with understanding and solving in a single inference. To examine this design choice, we first verify that the translation stage does not compromise performance. Out of the 8,011 questions in \dataset, we observe only 15 cases (0.18\%) where translation fails, all due to context length limitations. Notably, the translation stage serves purely as a user-friendly feature without affecting the model's problem-solving capabilities.

For generating Korean solutions, we compare three approaches: 
(1) direct generation in Korean (K2K), (2) our single-pass \method, and (3) multi-step inference (MSI). MSI is a direct re-implementation of \method through prompting that separately performs three steps: translating the Korean question to English, solving in English, and re-translating the solution back to Korean. In Table~\ref{tab_elo}, we evaluate these methods across three metrics: accuracy on \dataset, response quality via ELO ratings\footnote{ELO rating is a widely adopted metric for comparing relative quality between language models through pairwise comparisons.}, and computational efficiency through token consumption. While our model achieves accuracy comparable to MSI, it demonstrates superior response quality - preferred in 87.32\% of direct comparisons (excluding ties). Furthermore, our approach consumes only 66\% of the tokens required by MSI, making it computationally more efficient. Detailed evaluation methodology and prompts are provided in Appendices~\ref{app:eval_methods}, \ref{app:prompt_template}.

\begin{table*}[ht]
\centering
\fontsize{11}{13}\selectfont
\begin{tabular}{lccccc}
\toprule
\textbf{Models} & \textbf{HUMSS} & \textbf{STEM} & \textbf{Applied Science} & \textbf{Other} & \textbf{Avg.}\\
\midrule
Qwen2.5-7B & 37.3 & 45.0 & 42.4 & 36.5 & 40.3 \\
(Ours) & \textbf{39.3} & \textbf{49.5} & \textbf{45.3} & \textbf{40.3}  & \textbf{43.6} \\ 
\bottomrule
\end{tabular}
\caption{Evaluation results on KMMLU~\citep{son2024kmmlu}.}
\label{tab:generalization_kmmlu}
\end{table*}

\begin{table*}[ht]
\centering
\fontsize{11}{13}\selectfont
\begin{tabular}{lcccccccccccc}
\toprule
\textbf{Models} & \textbf{BN} & \textbf{DE} & \textbf{EN} & \textbf{ES} & \textbf{FR} & \textbf{JA} & \textbf{RU} & \textbf{SW} & \textbf{TE} & \textbf{TH} & \textbf{ZH} & \textbf{Av.} \\
\midrule
Qwen2.5-7B & 60.8 & 79.6 & \textbf{90.8} & 80.0 & 77.2 & 70.0 & \textbf{84.0} & \textbf{18.4} & 31.2 & \textbf{76.4} & \textbf{82.8} & 68.3 \\
(Ours) & \textbf{65.6} & \textbf{80.4} & 90.0 & \textbf{83.6} & \textbf{80.0} & \textbf{75.2} & 83.2 & 14.8 & \textbf{48.8} & 71.6 & 81.6 & \textbf{70.4} \\ 
\bottomrule
\end{tabular}
\caption{Evaluation results on MGSM~\citep{shi2022language}.}
\label{tab:generalization_mgsm}
\end{table*}

\subsection{Ablation Analysis}\label{sec_ust_ablation}

The \method dataset consists of three stages: Understand, Solve, and Translate. Having confirmed that the translation stage does not impact performance, we now examine the roles of the Understanding and Solving stages. We conduct experiments with four different configurations by varying both the presence and language of these stages. In our experiments, `-' indicates the omission of a stage, while `Korean' or `English' specifies the language used. For computational efficiency, we use a smaller model (Qwen2.5-1.5B-Instruct) and randomly sample 50k instances from the original dataset.

Our experiments reveal that the original configuration - both stages in English - achieves the highest performance (44.43\%) among all variants. A notable finding is that adding a Korean understanding stage actually decreases performance (36.49\% → 34.82\%). We attribute this counterintuitive result to two factors. First, when solving in Korean, an explicit understanding stage may be redundant as it essentially serves as another form of translation. Second, and more importantly, this suggests that chain-of-thought reasoning is most effective when conducted in the model's preferred language (English). This aligns with our observation that models show weaker reasoning capabilities in non-English languages, likely due to limited exposure during pre-training.

\section{Tracing the Performance Gains}

In this work, we demonstrate that routing through English understanding and reasoning steps enhances model performance on \dataset. To understand the source of these improvements, we evaluate our \traindataset-trained model on two additional benchmarks: KMMLU~\citep{son2024kmmlu} and MGSM~\citep{shi2022language}.

Our model shows consistent improvements across all KMMLU categories, with the largest gains in STEM (+4.5) and the smallest in HUMSS (+2.0)\footnote{This aligns with the nature of CoT, which primarily enhances reasoning capabilities~\citep{sprague2024cot} rather than factual knowledge required for HUMSS questions.}. However, when tested on MGSM, the model shows performance drops in Swahili (-3.6) and Thai (-4.8). These contrasting results suggest that our gains stem not from general improvements in mathematical reasoning, but rather from enhanced Korean-specific capabilities and better Korean-to-English reasoning alignment.

Our findings align with our initial goal: addressing the performance gap between English and Korean reasoning on identical questions (Section~\ref{sec_ust_performance}). The effectiveness of our approach is demonstrated in two ways. First, it recovers most of the performance achieved with E2E prompting (Table~\ref{tab:model_performance}). Second, it shows successful transfer to new domains (Table~\ref{tab:generalization_kmmlu}), suggesting that reasoning capabilities learned from machine-verifiable mathematics can generalize effectively. Most importantly, our method provides a simple path for non-English language users to benefit from the advanced reasoning capabilities typically available only in English.

\section{Conclusion}

In this paper, we propose \method, a training method that leverages English as an anchor language to enhance reasoning capabilities in Korean, and introduce \dataset, a benchmark of 8,011 English-Korean parallel mathematics problems. Our analysis reveals that the performance gap in multilingual reasoning primarily stems from difficulties in processing non-English inputs. Through extensive experiments, we demonstrate that \method effectively bridges this gap and shows promising generalization to various Korean domains beyond mathematics. Our approach offers a simple yet effective solution for non-English language users to benefit from advanced reasoning capabilities typically available only in English, suggesting a practical direction for improving multilingual reasoning capabilities in language models.


\newpage

\bibliography{custom}

\appendix
\section{Dataset Details}\label{app:detail_on_dataset}

This section provides additional details regarding the construction and composition of the \dataset benchmark.

\subsection{Dataset Sources}

\paragraph{KSM} 
This subset consists of problems from Korean mathematics examinations and competitions:

\begin{itemize}
    \item \textbf{College Scholastic Ability Test (CSAT)}\footnote{\url{https://www.suneung.re.kr/boardCnts/list.do?boardID=1500234&m=0403&s=suneung&searchStr=}}: The Korean counterpart to the SAT, which serves as the primary university entrance examination. We include only math problems with historical error rates exceeding 70\%.
    \item \textbf{Korean Mathematical Olympiad (KMO)}\footnote{\url{https://www.kmo.or.kr/kmo/sub07.html}}: The Korean equivalent to the International Mathematical Olympiad (IMO), primarily designed for middle- and high-school students. These problems require advanced mathematical knowledge and critical thinking. 
    \item \textbf{Korean Junior Mathematical Olympiad (KJMO)}\footnote{\url{https://www.kms.or.kr/board/list.html?code=junior2}}: An elementary-school version of KMO aimed at identifying mathematical talent at an early stage.
    \item \textbf{Korean University Mathematical Olympiad (KMS)}\footnote{\url{https://www.kms.or.kr/board/list.html?code=conf12}}: A university-level competition featuring advanced topics in calculus, linear algebra, number theory, geometry, and discrete mathematics.
    \item \textbf{Korean National Teacher Qualification Test (TQ)}\footnote{\url{https://blog.naver.com/headracer}}: A standardized examination for teacher certification that focuses on mathematical pedagogy and content expertise.
\end{itemize}

\paragraph{Prior Sets} 
This subset incorporates problems drawn from established English mathematics benchmarks, filtered to include only those with numeric answers:

\begin{itemize}
    \item \textbf{GSM8K}~\citep{cobbe2021training}: A collection of 8.5K grade-school math word problems that emphasize multi-step reasoning and elementary arithmetic.
    \item \textbf{MATH}~\citep{hendrycks2020measuring}: A comprehensive benchmark of 12.5K high-school competition-level problems spanning seven mathematical domains, each accompanied by detailed step-by-step solutions.
    \item \textbf{Omni-MATH}~\citep{gao2024omni}: An advanced dataset containing 4.4K Olympiad-level problems across 33 sub-domains and 10 difficulty tiers, designed to push the limits of current LLM capabilities.
    \item \textbf{MMMLU}~\citep{openai_mmmlu}: A multilingual extension of the MMLU benchmark~\citep{hendrycks2020measuring}, covering various STEM fields (e.g., abstract algebra, college mathematics, and high-school mathematics). It is available in 14 languages produced by professional translators.
\end{itemize}

\subsection{Post-processing}

\begin{figure*}
    \centering
    \includegraphics[width=\linewidth]{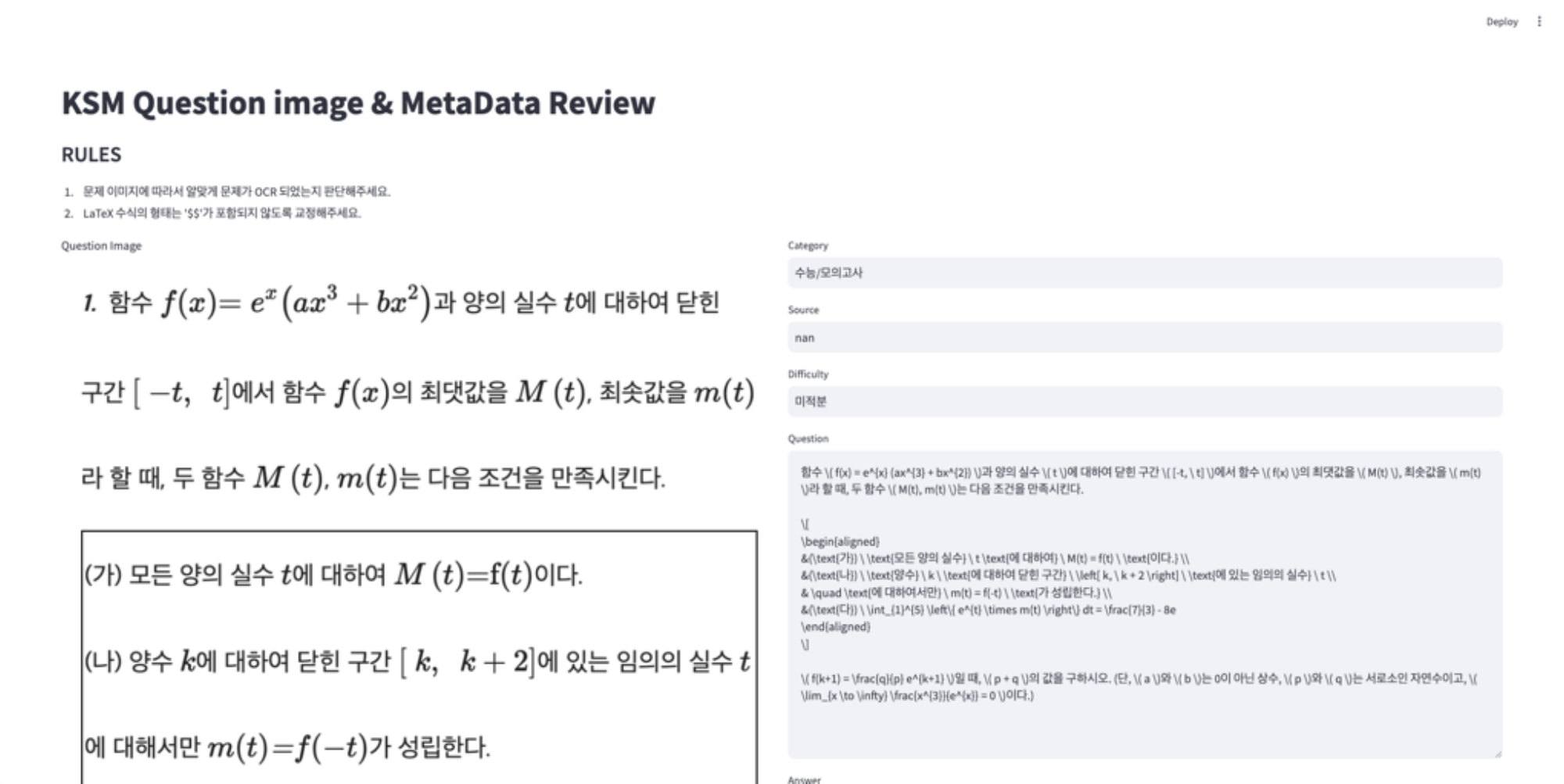}
    \caption{Screenshot of our Streamlit-based OCR validation tool, used to compare source documents with OCR outputs and correct any errors. }
    \label{fig:streamlit_review}
\end{figure*}

For the \textbf{KSM} subset, we performed a manual verification and editing procedure to ensure high-quality OCR results. Specifically, we developed a review application in Streamlit\footnote{\url{https://streamlit.io/}}, illustrated in Figure~\ref{fig:streamlit_review}, which compares the original problem text against the OCR output. Two main factors were verified:

\begin{itemize}
    \item \emph{Content Completeness}: Confirming that all parts of the problem statement are accurately captured and that no text is omitted.
    \item \emph{\LaTeX~Integrity}: Ensuring that mathematical symbols and equations are correctly transcribed in \LaTeX\ format.
\end{itemize}

Based on these checks, we corrected errors and added any missing content. For instance, monetary symbols (\$) enclosing \LaTeX~symbols were removed to enhance clarity. Erroneous OCR outputs were manually fixed, and missing text was supplemented as needed. Figure~\ref{fig:streamlit_review} illustrates our interactive review interface.

\section{Additional Details in Experiments}\label{app:detail_on_exp}

We experiment with six multilingual language models reported to have been pretrained on Korean data: three Qwen2.5 Instruct models (1.5B, 7B, and 72B parameters)~\citep{Qwen2024} and three Llama-3.1/2 Instruct models (1B, 8B, and 70B parameters)~\citep{dubey2024llama, Meta2024llama32}. For simplicity, we omit the word ``Instruct'' in references to these models, although all are instruction-tuned.

Unless otherwise noted, we set the sampling temperature to 0.7 and top\_p to 0.95, with a minimum of 8 tokens and maximum of 2{,}048 tokens for the output. While lower temperatures are often used in pass@1 settings, we observed that extremely low temperatures sometimes cause models to revert to their preferred language (often English or Chinese). Hence, to maintain the specified response language, we employ a slightly higher temperature with moderate top\_p.

\section{Fine-tuning Details}\label{app:finetuning_details}

We fine-tune our models on H100 80GB GPUs using DeepSpeed ZeRO. Specifically, we train Qwen2.5-7B-Instruct on the \traindataset approach and conduct ablation analyses with Qwen2.5-1.5B-Instruct under various settings. To maximize GPU utilization, we use a batch size of 96 per GPU across four GPUs for Qwen2.5-7B-Instruct, and a batch size of 128 per GPU across two GPUs for Qwen2.5-1.5B-Instruct. Table~\ref{tab:hyperparameters} summarizes the relevant hyperparameters.

\begin{table*}[ht]
\centering
\caption{Hyperparameters for fine-tuning and ablation studies.}
\fontsize{10}{12}\selectfont
\begin{tabular}{cccccccc}
\toprule
\textbf{Base Model} & \textbf{Batch Size} & \textbf{Learning Rate} & \textbf{Scheduler} & \textbf{Optimizer} & \textbf{Max Length} & \textbf{\# GPUs} \\
\midrule
Qwen2.5-7B & 96 & 2e-5 & Cosine & AdamW & 8192 & 4 \\
Qwen2.5-1.5B & 128 & 2e-5 & Cosine & AdamW & 8192 & 2 \\
\bottomrule
\end{tabular}
\label{tab:hyperparameters}
\end{table*}

\section{Multi-Step Inference}\label{app:msi}

Multi-Step Inference (M.S.I) is a direct re-implementation of \method solely via prompting (i.e., without additional training). It is carried out in three steps:

\begin{enumerate}
    \item \textbf{Translated-English (TE)}: Translate the original Korean prompt into English. We use the template shown in Figure~\ref{fig:te_trans}.
    \item \textbf{Translated-English-to-English (TE2E)}: Solve the translated problem in English using the template in Figure~\ref{fig:solution_en_reasoning}.
    \item \textbf{Translated-English-to-English-to-Korean (TE2E2K)}: Translate the English solution back into Korean, following the template in Figure~\ref{fig:te2e_solution_trans}.
\end{enumerate}

This multi-step approach echoes the three-stage \method pipeline (Understand, Solve, Translate) but relies on separate inferences with task-specific prompts.

\section{Evaluation Methods}\label{app:eval_methods}

\subsection{ELO Rating}

We use an Elo rating system to compare responses produced by different approaches in a pairwise manner~\citep{lmsys2024chat}. Elo ratings are computed in two parts: (1) \emph{Expected Score}, which gauges each model’s probability of winning based on current ratings; and (2) \emph{Rating Update}, which adjusts the ratings after each match.

\paragraph{Expected Score.}
Given two models \(A\) and \(B\) with Elo ratings \(R_{A}\) and \(R_{B}\), their expected scores \(E_A\) and \(E_B\) are:

\begin{align*}
    E_{A} = \frac{1}{1 + 10^{\left(R_{B} - R_{A}\right)/400}}, \\ \quad
    E_{B} = \frac{1}{1 + 10^{\left(R_{A} - R_{B}\right)/400}}.
\end{align*}

\paragraph{Rating Update.}
After each pairwise comparison, the rating of model \(A\) is updated as follows:
\begin{align*}
    R'_{A} = R_{A} + K \times (S_{A} - E_{A}),
\end{align*}

where \(S_A \in \{0,1\}\) is the actual score (\(1\) if \(A\) is preferred, and \(0\) otherwise). The constant \(K\) modulates the step size of the rating update; we set \(K=4\) for more stable ratings. We also randomly shuffle match order and apply bootstrapping over 1{,}000 iterations to mitigate dependence on match sequence~\citep{boubdir2023elo}.

\subsection{Token Consumption}

We measure token consumption via a simplified metric that accounts for both input and output tokens. For a dataset of \(N\) samples, let \(T_{\text{input}}\) and \(T_{\text{output}}\) be the number of input and output tokens, respectively, for each sample. The total token cost \(L_{\text{model}}\) for each model is:

\[
L_{\text{model}} = \frac{\sum_{i=1}^{N}\left(T_{i,\text{input}} + 3 \times T_{i,\text{output}}\right)}{N},
\]

where we weight output tokens by a factor of 3 to reflect their higher processing cost, following cost ratios from common LLM providers (e.g., OpenAI, Mistral AI, Alibaba Cloud, and Deepseek AI).




\begin{table*}[t]
\centering
\caption{Evaluation result for Qwen and Llama models on \dataset. All models are instruction-tuned, but they are abbreviated for simplicity.}
\fontsize{11}{13}\selectfont
\begin{tabular}{lcccccc}
\toprule
\multicolumn{1}{c}{\textbf{Model}} & \textbf{GSM8K} & \textbf{MATH} & \textbf{Omni-MATH} & \textbf{MMMLU} & \textbf{KSM} & \textbf{Avg.} \\
\midrule
\multicolumn{7}{c}{\textit{\textbf{Korean-to-Korean (K2K)}}} \\ 
\midrule
Qwen2.5-1.5B & 28.13 & 20.69 & 8.64 & 18.51 & 3.78 & 15.95 \\
Qwen2.5-7B & 66.41 & 50.36 & 18.96 & 50.00 & 11.83 & 39.52 \\
Qwen2.5-72B & 89.46 & 74.73 & 30.07 & 69.79 & 25.35 & 57.88 \\
Llama-3.2-1B & 7.88 & 10.50 & 4.77 & 10.43 & 2.80 & 7.28 \\
Llama-3.1-8B & 57.47 & 31.20 & 11.73 & 32.98 & 5.25 & 27.73 \\
Llama-3.1-70B & 78.62 & 56.12 & 20.38 & 57.87 & 11.83 & 44.96 \\
\midrule
\multicolumn{7}{c}{\textit{\textbf{Korean-to-English (K2E)}}} \\ \midrule
Qwen2.5-1.5B & 31.92 & 26.86 & 10.32 & 30.85 & 5.81 & 21.15 \\
Qwen2.5-7B & 65.20 & 54.59 & 20.22 & 49.79 & 16.67 & 41.29 \\
Qwen2.5-72B & 89.23 & 77.68 & 32.74 & 70.43 & 27.73 & 59.56 \\
Llama-3.2-1B & 7.20 & 9.57 & 5.08 & 11.49 & 2.94 & 7.26 \\
Llama-3.1-8B & 55.04 & 31.13 & 11.16 & 27.66 & 5.18 & 26.03 \\
Llama-3.1-70B & 77.63 & 55.18 & 19.43 & 58.94 & 12.82 & 44.80 \\
\midrule
\multicolumn{7}{c}{\textit{\textbf{English-to-English (E2E)}}} \\ \midrule
Qwen2.5-1.5B & 65.50 & 49.25 & 16.87 & 45.53 & 8.33 & 37.10 \\
Qwen2.5-7B & 81.35 & 68.87 & 27.29 & 57.02 & 21.08 & 51.12 \\
Qwen2.5-72B & 94.31 & 83.33 & 37.72 & 70.00 & 31.65 & 63.40 \\
Llama-3.2-1B & 43.44 & 27.24 & 9.90 & 23.83 & 3.64 & 21.61 \\
Llama-3.1-8B & 79.45 & 48.11 & 16.08 & 42.34 & 7.21 & 38.64 \\
Llama-3.1-70B & 93.33 & 67.90 & 24.83 & 70.43 & 17.09 & 54.71 \\
\bottomrule
\end{tabular}
\label{tab:model_performance}
\end{table*}

\begin{table*}[t]
\centering
\caption{Translated-English-to-English (TE2E) prompting evaluation result on \dataset.}
\fontsize{11}{13}\selectfont
\begin{tabular}{lcccccc}
\toprule
\multicolumn{1}{c}{\textbf{Model}} & \textbf{GSM8K} & \textbf{MATH} & \textbf{Omni-MATH} & \textbf{MMMLU} & \textbf{KSM} & \textbf{Avg.} \\
\midrule
\multicolumn{7}{c}{\textit{\textbf{Translated-English-to-English (TE2E)}}} \\ \midrule
Qwen2.5-1.5B & 36.24 & 34.52 & 12.05 & 29.36 & 6.51 & 23.74 \\
Qwen2.5-7B & 79.53 & 70.78 & 28.86 & 59.36 & 20.38 & 51.78 \\
Qwen2.5-72B & 89.16 & 78.13 & 34.05 & 70.43 & 31.65 & 60.68 \\
Llama-3.2-1B & 2.96 & 5.41 & 1.94 & 9.36 & 1.75 & 4.28 \\
Llama-3.1-8B & 53.90 & 36.43 & 13.04 & 29.57 & 4.76 & 27.54 \\
Llama-3.1-70B & 78.17 & 60.83 & 22.26 & 60.43 & 16.25 & 47.59 \\
\bottomrule
\end{tabular}
\label{tab:translation_performance}
\end{table*}

\section{Cross-Lingual Application of RMs}\label{app:rm_application}

\begin{figure}[h]
    \centering
    \includegraphics[width=\columnwidth]{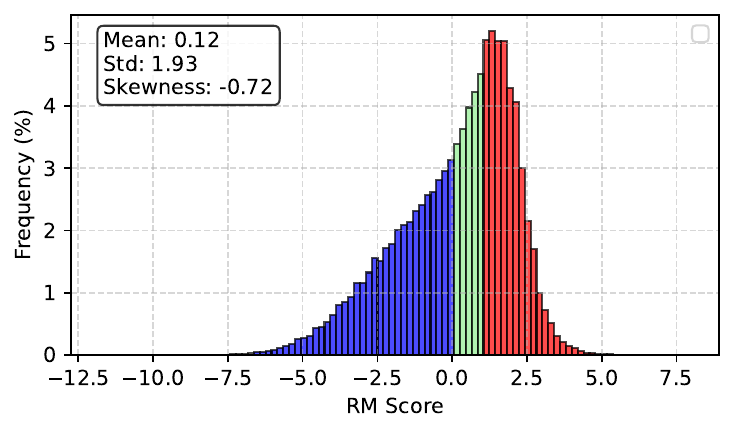}
    \caption{\footnotesize Reward model evaluation result on \method dataset. The samples were categorized into three groups based on the reward model score: high (RM Score $>$ 1, red), low (RM Score $<$ 0, blue), and medium (0 $\leq$ RM Score $\leq$ 1, green).}
    \label{fig_rm_score}
\end{figure}

While creating the \traindataset we leverage Qwen2.5-Math-RM-72B a reward model (RM) originally intended to be used in English or Chinese. We observe whether such RMs can be applied with further post-training for language transfer. In Figure~\ref{fig_rm_score}, we illustrate the score distribution on our initial dataset. The distribution shows to be right-skewed with a gradual tapering off towards the left. We create two datasets high and low. The high consists of samples with a score higher than 1 (colored in red) and low consists samples with a score lower than 0 (colored in blue). The high is used as our final dataset. For comparison we train a Qwen2.5-7B-Instruct model on the low dataset with identical number of instances. 

\begin{figure}[h]
    \centering
    \includegraphics[width=\columnwidth]{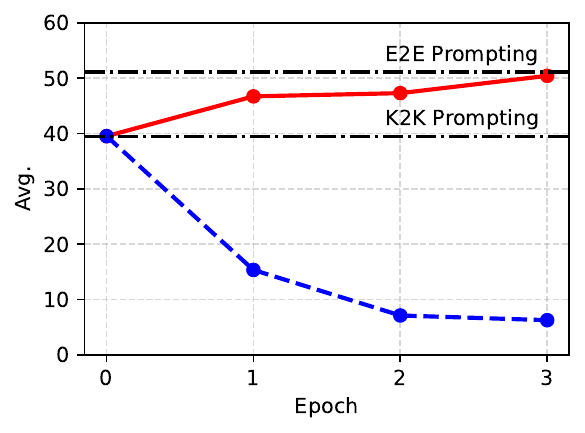}
    \caption{\footnotesize Qwen2.5-7B-Instruct model performance trends across epochs during training on high (red) and low (blue) datasets. The evaluation results of the original Qwen2.5-7B-Instruct model for K2K and E2E prompting were depicted with the dash-dotted lines.}
    \label{fig_rm_results}
\end{figure}

As shown in Figure~\ref{fig_rm_results}, training on the High subset progressively improves the model's performance. In contrast, using the Low subset degrades performance, with the model's score dropping to an average of 6.25 on \dataset. This result suggests that RMs can be applied to new languages without additional training, aligning with previous studies~\citep{son2024llm, son2024mm}. Accordingly, we choose to use the filtered subset.

\section{Prompt Templates}\label{app:prompt_template}

This section provides the complete text of all prompts used in our experiments, evaluations, and dataset construction. Each prompt is presented in a separate figure, preserving the original structure while enhancing clarity and consistency.  

\begin{enumerate}
    \item \textbf{OCR Prompt (Figure~\ref{fig:ocr_prompt})}: Performs OCR on an image of a Korean math problem and extracts the text. Used to build the KSM subset (Section~\ref{sec_bencmark_formulation}).
    \item \textbf{Question Translation Prompts (Figures~\ref{fig:en_ko_questions_trans}--\ref{fig:ko_en_question_trans})}: Translate math questions into Korean or English, respectively. Used for creating bilingual pairs in \dataset (Section~\ref{sec_construction}).
    \item \textbf{Solution Generation Prompts (Figures~\ref{fig:solution_en_reasoning}--\ref{fig:solution_ko_reasoning})}: Evaluate model performance on \dataset under different reasoning-language conditions (Section~\ref{sec_experiment_design}).
    \item \textbf{Understanding Generation Prompt (Figure~\ref{fig:understanding_generation})}: Produces a structured breakdown of the problem for the \textit{Understanding} stage of \method (Section~\ref{sec_ust_construction}).
    \item \textbf{Model Translation Prompts (Figures~\ref{fig:te_trans}--\ref{fig:te2e_solution_trans})}: Used in multi-step inference (M.S.I) to translate the question or solution between Korean and English (Section~\ref{sec_multi_task}).
    \item \textbf{LLM-as-a-Judge Prompt (Figure~\ref{fig:llm_as_a_judge})}: Conducts pairwise response comparisons and produces a verdict, enabling ELO-based evaluation (Section~\ref{sec_ust_performance}).
\end{enumerate}

\begin{figure}[h!]
\fontsize{8}{9}\selectfont
    \centering
    \begin{tabular}{|p{0.9\linewidth}|}
    \hline 
    \\
    \textbf{OCR Prompt} \\
        \\
        You will be given an image containing a Korea math question. Your task is to conduct an OCR to retrieve the question in text format. \\
        \\
        Follow the following roles: \\
        \\
        1. return the question only, nothing else. \\
        2. If, the image contains the answer ignore it. Do not return it with the question. \\
        3. Put extra care on notations and equations make sure they are identical. \\
\\\\
\hline
    \end{tabular}
    \caption{Prompt to perform OCR and extract the mathematical question from the given image. Both the prompt and the screenshot of the problem were provided to the model for OCR processing.}
    \label{fig:ocr_prompt}
\end{figure}

\begin{figure}[h!]
\fontsize{8}{9}\selectfont
    \centering
    \begin{tabular}{|p{0.9\linewidth}|}
    \hline 
    \\
    \textbf{Question Translation Prompt: En\(\to\)Ko} \\
        \\
        You are a professional English-to-Korean translator specializing in academic content. Your task is to translate math problems provided in English into clear, natural, and precise Korean referring to given examples. Follow the instructions below: \\
        \\
        \#\#\# INSTRUCTIONS: \\
        1. You SHOULD NOT solve the problem and translate only the given question — do not include any additional commentary. \\
        2. Preserve all mathematical symbols, notations, formatting, and existing choices exactly as presented. \\
        3. Use fluent, natural Korean that aligns with academic standards for math problems. \\
        4. Ensure the translation conveys the meaning and context accurately. \\
        \\
        \#\#\# INPUT: \\
        \{question\}
\\\\
\hline
    \end{tabular}
    \caption{Translation prompt to translate English math questions into Korean. This prompt is utilized to translate the \textit{Prior Sets} data, sourced from English math benchmarks, into Korean. The bracketed part is a placeholder to fill in the question.}
    \label{fig:en_ko_questions_trans}
\end{figure}

\begin{figure}[h!]
\fontsize{8}{9}\selectfont
    \centering
    \begin{tabular}{|p{0.9\linewidth}|}
    \hline 
    \\
    \textbf{Question Translation Prompt: Ko\(\to\)En} \\
        \\
        You are a professional Korean-to-English translator specializing in academic content. Your task is to translate math problems provided in Korean into clear, natural, and precise English referring to given examples. Follow the instructions below: \\
        \\
        \#\#\# INSTRUCTIONS: \\
        1. You SHOULD NOT solve the problem and translate only the given question — do not include any additional commentary. \\
        2. Preserve all mathematical symbols, notations, formatting, and existing choices exactly as presented. \\
        3. Use fluent, natural English that aligns with academic standards for math problems. \\
        4. Ensure the translation conveys the meaning and context accurately. \\
\\\\
\hline
    \end{tabular}
    \caption{Translation prompt to translate Korean math questions into English. This prompt is utilized to translate the \textit{KSM} data, sourced from Korean math examinations and competitions, into English. The bracketed part is a placeholder to fill in the question.}
    \label{fig:ko_en_question_trans}
\end{figure}

\begin{figure}[ht!]
\fontsize{8}{9}\selectfont
    \centering
    \begin{tabular}{|p{0.9\linewidth}|}
    \hline 
    \\
    \textbf{\dataset Solution Prompt: English Reasoning} \\
        \\
        Solve the given question. \\
        After solving the problem, state your final answer in the following format: \$\textbackslash\textbackslash boxed\{N\}\$. \\
        \\
        \{question\} Respond in English.\\
\\\\
\hline
    \end{tabular}
    \caption{Solution generation prompt to evaluate the models in an English reasoning setup, such as K2E and E2E. The bracketed part is a placeholder to fill in the question.}
    \label{fig:solution_en_reasoning}
\end{figure}

\begin{figure}[ht!]
\fontsize{8}{9}\selectfont
    \centering
    \begin{tabular}{|p{0.9\linewidth}|}
    \hline 
    \\
    \textbf{\dataset Solution Prompt: English Reasoning} \\
        \\
        Solve the given question. \\
        After solving the problem, state your final answer in the following format: \$\textbackslash \textbackslash boxed\{N\}\$. \\
        \\
        \{question\} Respond in Korean.\\
\\\\
\hline
    \end{tabular}
    \caption{Solution generation prompt to evaluate the models in a Korean reasoning setup, such as K2K. The bracketed part is a placeholder to fill in the question.}
    \label{fig:solution_ko_reasoning}
\end{figure}

\begin{figure}[h!]
\fontsize{8}{9}\selectfont
    \centering
    \begin{tabular}{|p{0.9\linewidth}|}
    \hline 
    \\
    \textbf{Understanding Generation Prompt} \\
        \\
        \text{[User]} \\
        Solve the following problem: \\
        \{question\} \\
        \\
        \text{[Assistant]} \\
        \{solution\} \\
        \\
        \text{[User]} \\
        I'm planning to generate a step-by-step guide for the solution. The step-by-step solution will be provided to students to guide their solution. Accordingly, it should be clear and straightforward, guiding the student through the problem-solving process. However, it must not reveal the answer as it will disturb the students' solution. Generate the breakdown. It should assist with understanding the question and planning how to solve it. The generation will be directly provided to the student, accordingly do not include notes like 'not reveal the answer', or a evaluation of your own breakdown. Write in first person view: e.g I will~, I can~. \\
\\\\
\hline
    \end{tabular}
    \caption{Generation prompt to generate \textit{Understanding} stage of \method. Given a problem and its corresponding answer, a prefix conversation history is constructed where the user asks the problem and the assistant provides the ground-truth answer. Subsequently, the user instructs the assistant to generate an understanding of the problem. The bracketed parts are a placeholder to fill in the question and its ground-truth solution.}
    \label{fig:understanding_generation}
\end{figure}

\begin{figure}[ht!]
\fontsize{8}{9}\selectfont
    \centering
    \begin{tabular}{|p{0.9\linewidth}|}
    \hline 
    \\
    \textbf{Model Translation Prompt: Translated-English (TE)} \\
        \\
        You are a professional Korean-to-English translator specializing in academic content. Your task is to translate math problems provided in Korean into clear, natural, and precise English referring to given examples. Follow the instructions below: \\
        \\
        \#\#\# INSTRUCTIONS: \\
        1. You SHOULD NOT solve the problem and translate only the given question — do not include any additional commentary. \\
        2. Preserve all mathematical symbols, notations, formatting, and existing choices exactly as presented. \\
        3. Use fluent, natural English that aligns with academic standards for math problems. \\
        4. Ensure the translation conveys the meaning and context accurately. \\
        \\
        \#\#\# INPUT: \\
        \text{[1st Korean Question Example]} \\
        \\
        \#\#\# OUTPUT: \\
        \text{[1st English Translation Result]} \\
        \\
        \(\cdots\) \\
        \\
        \#\#\# INPUT: \\
        \text{[5th Korean Question Example]} \\
        \\
        \#\#\# OUTPUT: \\
        \text{[5th English Translation Result]} \\
        \\
        \#\#\# INPUT: \\
        \{question\} \\
        \\
        \#\#\# OUTPUT:
\\\\
\hline
    \end{tabular}
    \caption{Translation prompt to generate Translated-English (TE) problem by translating the given Korean problem into English. The bracketed part is a placeholder to fill in the question.}
    \label{fig:te_trans}
\end{figure}

\begin{figure}[t]
\fontsize{8}{9}\selectfont
    \centering
    \begin{tabular}{|p{0.9\linewidth}|}
    \hline 
    \\
    \textbf{Model Translation Prompt: Translated-English-to-English-to-Korean (TE2E2K)} \\
        \\
        You are a professional Korean-to-English translator specializing in academic content. Your task is to translate math problems provided in Korean into clear, natural, and precise English referring to given examples. Follow the instructions below: \\
        \\
        \#\#\# INSTRUCTIONS: \\
        1. You SHOULD NOT solve the problem and translate only the given question — do not include any additional commentary. \\
        2. Preserve all mathematical symbols, notations, formatting, and existing choices exactly as presented. \\
        3. Use fluent, natural English that aligns with academic standards for math problems. \\
        4. Ensure the translation conveys the meaning and context accurately. \\
        \\
        \#\#\# INPUT: \\
        \text{[1st English Solution Example]} \\
        \\
        \#\#\# OUTPUT: \\
        \text{[1st Korean Translation Result]} \\
        \\
        \(\cdots\) \\
        \\
        \#\#\# INPUT: \\
        \text[{5th English Solution Example]} \\
        \\
        \#\#\# OUTPUT: \\
        \text{[5th Korean Translation Result]} \\
        \\
        \#\#\# INPUT: \\
        \{question\} \\
        \\
        \#\#\# OUTPUT:
\\\\
\hline
    \end{tabular}
    \caption{Translation prompt to generate Translated-English-to-English-to-Korean solution by translating the given English solution into Korean. The bracketed part is a placeholder to fill in the question.}
    \label{fig:te2e_solution_trans}
\end{figure}

\begin{figure}[t]
\fontsize{8}{9}\selectfont
    \centering
    \begin{tabular}{|p{0.9\linewidth}|}
    \hline 
    \\
    \textbf{LLM-as-a-judge Prompt} \\
        \\
        \text{[System]} \\
        Please act as an impartial judge and evaluate the quality of the responses provided by two AI assistants to the user question displayed below. You should choose the assistant that follows the user’s instructions and answers the user’s question better. Your evaluation should consider factors such as the helpfulness, relevance, accuracy, depth, creativity, and level of detail of their responses. Begin your evaluation by comparing the two responses and provide a short explanation. Avoid any position biases and ensure that the order in which the responses were presented does not influence your decision. Do not allow the length of the responses to influence your evaluation. Do not favor certain names of the assistants. Be as objective as possible. After providing your explanation, output your final verdict by strictly following this format: "[[A]]" if assistant A is better, "[[B]]" if assistant B is better. There is no option for a tie, you should choose "[[A]]" or "[[B]]". \\
        \\
        \text{[User Question]} \\
        \{question\} \\
        \\
        \text{[The Start of Assistant A’s Answer]} \\
        \{model\_a\_answer\} \\
        \text{[The End of Assistant A’s Answer]} \\
        \\
        \text{[The Start of Assistant B’s Answer]} \\
        \{model\_b\_answer\} \\
        \text{[The End of Assistant B’s Answer]} \\
\\\\
\hline
    \end{tabular}
    \caption{Evaluation prompt to conduct llm-as-a-judge pairwise evaluation between model A's response and model B's response to a given question. The bracketed parts are the placeholders to fill in the question, model A's answer, and model B's answer.}
    \label{fig:llm_as_a_judge}
\end{figure}

\end{document}